\documentclass[conference]{IEEEtran}
\def\IEEEtitletopspace{18pt}
\IEEEoverridecommandlockouts
\usepackage{cite}
\usepackage{amsmath,amssymb,amsfonts}
\usepackage{algorithmic}
\usepackage{graphicx}
\usepackage{textcomp}
\usepackage{xcolor}
\usepackage{svg}
\usepackage{array}
\usepackage{siunitx}
\graphicspath{{./imgs/}}
\svgpath{{../imgs/}}
\def\BibTeX{{\rm B\kern-.05em{\sc i\kern-.025em b}\kern-.08em
  T\kern-.1667em\lower.7ex\hbox{E}\kern-.125emX}}
  
\newbox{\myorcidaffilbox}
\sbox{\myorcidaffilbox}{\large\includegraphics[height=1.25ex]{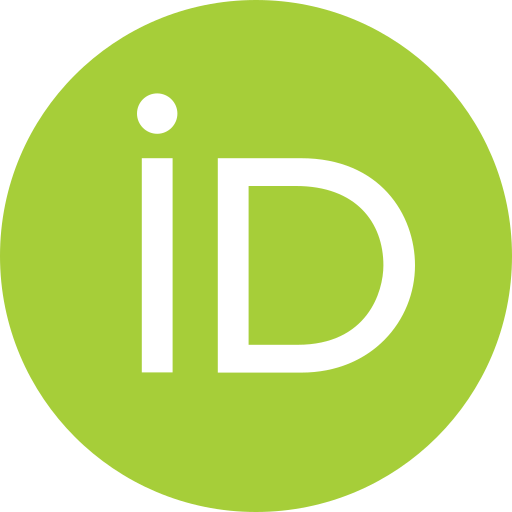}}
\newcommand{\orcidaffil}[1]{%
\href{https://orcid.org/#1}{\usebox{\myorcidaffilbox}}}
\usepackage{hyperref}

\begin{document}
\title{A Bioinspired Synthetic Nervous System Controller for Pick-and-Place Manipulation*\\
}

\author{
 Yanjun Li$^{1}$$^{\dagger}$\textsuperscript{\orcidaffil{0000-0002-2534-7049}},
 Ravesh Sukhnandan$^{2}$$^{\dagger}$\textsuperscript{\orcidaffil{0000-0001-5858-9610}}, Jeffrey P. Gill$^{3}$\textsuperscript{\orcidaffil{0000-0002-4115-8045}}, Hillel J. Chiel$^{3,4,5}$\textsuperscript{\orcidaffil{0000-0002-1750-8500}},
 Victoria Webster-Wood$^{2}$$^{+}$\textsuperscript{\orcidaffil{0000-0001-6638-2687}},\\ and Roger D. Quinn$^{1}$\textsuperscript{\orcidaffil{0000-0002-8504-7160}}
 
\vspace{-15pt}

\thanks{*This work was supported in part by the National Science Foundation (NSF) grant no. FRR-2138873 and by a GEM fellowship. Any opinions, findings, and conclusions or recommendations expressed in this material are those of the authors and do not necessarily reflect the views of the NSF.}
\thanks{\copyright 2023 IEEE.  Personal use of this material is permitted.  Permission from IEEE must be obtained for all other uses, in any current or future media, including reprinting/republishing this material for advertising or promotional purposes, creating new collective works, for resale or redistribution to servers or lists, or reuse of any copyrighted component of this work in other works.}
\thanks{$^{1}$\textit{Department of Mechanical Engineering, Case Western Reserve University, 10900 Euclid Ave, Cleveland, OH 44106, United States}}

\thanks{$^{2}$\textit{Department of Mechanical Engineering, Carnegie Mellon University, 5000 Forbes Ave, Pittsburgh, PA 15213, United States}}

\thanks{$^{3}$\textit{Department of Biology, Case Western Reserve University, 10900 Euclid Ave, Cleveland, OH 44106, United States}}

\thanks{$^{4}$\textit{Department of Neurosciences, Case Western Reserve University, 10900 Euclid Ave, Cleveland, OH 44106, United States}}

\thanks{$^{5}$\textit{Department of Biomedical Engineering, Case Western Reserve University, 10900 Euclid Ave, Cleveland, OH 44106, United States}}

\thanks{$^{\dagger}$These authors contributed equally to the work}

\thanks{$^{+}$Corresponding author
  {\tt\small vwebster@andrew.cmu.edu}}

  }


\maketitle

\begin{abstract}
The Synthetic Nervous System (SNS) is a biologically inspired neural network (NN). Due to its capability of capturing complex mechanisms underlying neural computation, an SNS model is a candidate for building compact and interpretable NN controllers for robots. Previous work on SNSs has focused on applying the model to the control of legged robots and the design of functional subnetworks (FSNs) to realize dynamical systems. However, the FSN approach has previously relied on the analytical solution of the governing equations, which is difficult for designing more complex NN controllers. Incorporating plasticity into SNSs and using learning algorithms to tune the parameters offers a promising solution for systematic design in this situation. In this paper, we theoretically analyze the computational advantages of SNSs compared with other classical artificial neural networks. We then use learning algorithms to develop compact subnetworks for implementing addition, subtraction, division, and multiplication. We also combine the learning-based methodology with a bioinspired architecture to design an interpretable SNS for the pick-and-place control of a simulated gantry system. Finally, we show that the SNS controller is successfully transferred to a real-world robotic platform without further tuning of the parameters, verifying the effectiveness of our approach.
\end{abstract}

\vspace{-6pt}
\section{Introduction}
\vspace{-5pt}


Animals are capable of remarkable behavioral diversity, including locomotion and dexterous manipulation, that results from the coupling of their neural and motor systems \cite{lyttle_robustness_2017,nishikawa_neuromechanics_2007}. In the pursuit of robotic control methodologies that can capture this behavioral flexibility, various biologically inspired neural controllers have been proposed at different levels of biological realism \cite{bekey_biologically_1996,vergara_villegas_advanced_2018,li_bio-inspired_2021,NCP_2}. Synthetic Nervous Systems (SNSs) are a promising approach to the control of robotic systems because (1) they incorporate elements of real neural dynamics, (2) the networks can be designed in a compact way to perform basic arithmetic operations, and (3) they have been used for real-time control of robots \cite{szczecinski_template_2017,szczecinski2017functional,hunt_development_2017, szczecinski2017design}.

SNSs have primarily been used in the past for the locomotion control in legged robots\cite{szczecinski_template_2017,hunt_development_2017,rubeo_synthetic_2018}. For example, the SNS approach has been used to successfully produce adaptive locomotion of the hind-legs of a dog-like robot \cite{hunt_development_2017}. SNS networks have also been used in the real-time control of the legs of the hexapod MantisBot for locomotion and steering \cite{szczecinski_template_2017}. However, SNS networks have not been previously demonstrated for grasping and manipulation control.

When designing complex controller architectures, one challenge for SNS networks is the number of parameters that must be tuned for stable performance. The Functional Sub-Network (FSN) approach has been used previously to tune the parameters of such SNS controllers by providing analytic constraints on the parameters of the arithmetic operations [8]. However, this approach requires analytical models of the subnetworks to perform optimization. To date, there has not been a systematic methodology that allows SNS network parameters to be learned for generic robotic motion control without using a closed-form analytical expression. 

In this work, we address these gaps by demonstrating the design and implementation of a learning-based SNS grasping controller in a pick-and-place task. We (1) present a learning-based methodology for tuning the parameters of the SNS, (2) present a biologically inspired architecture for the development of SNS controllers for real-time robotic control, (3) expand the use of SNS controllers beyond locomotion and into the realm of real-time manipulation tasks, and (4) show that the SNS controller can be successfully trained in simulation and deployed on a real-world XYZ grasping system without additional parameter tuning.

\vspace{-4pt}

\section{Methodology}
\vspace{-2pt}

\subsection{Discretization of Synthetic Nervous Systems}\label{DSNS}

\begin{figure}
 \centering
 \includegraphics[width=0.94\columnwidth]{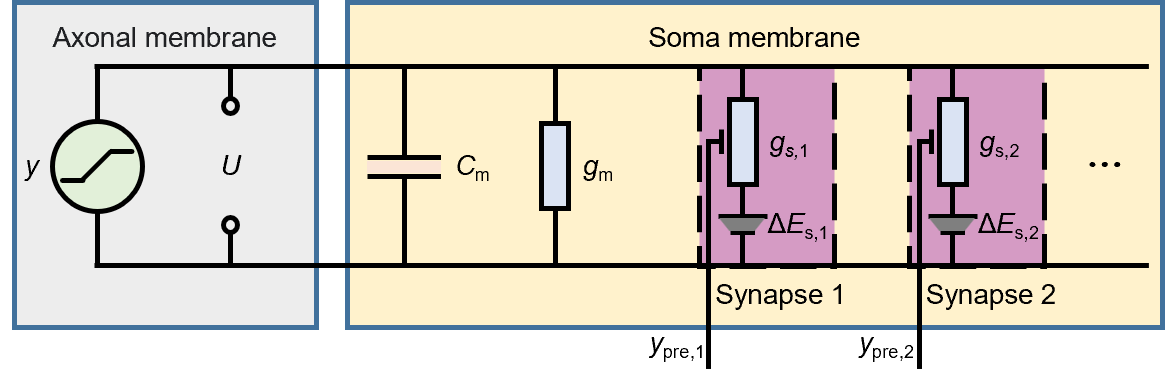}
 \vspace{-13pt}
 \caption{Schematic of Synthetic Nervous System components. The SNS model incorporates membrane capacitance, leak conductance, and synaptic channels. The conductance currents govern the evolution of the membrane potential $U$. Instead of utilizing voltage-gated ion channels to explicitly generate spikes, the SNS model uses neural activity $y$ to reflect temporal firing frequency.}
 \vspace{-15pt}
 \label{fig:SNS}
\end{figure}

A Synthetic Nervous System (SNS) is a neural network (NN) inspired by neuroscience (Fig.~\ref{fig:SNS}). It considers synapses as ionic channels with different reversal potentials \cite{10.1093/oso/9780195104912.001.0001}. The synaptic strength can be adjusted by changing the conductance. The governing equation of the $i$th neuron within an SNS can be written as the following ordinary differential equation:
\begin{equation}\label{eq:1}
  C_{\text{m},i}\frac{\mathrm{d}U_i}{\mathrm{d}t} = g_{\text{m},i}(E_{\text{r},i} - U_i) + \sum\limits_{j} g_{ij}y_j \left(E_{ij} - U_i\right) + I_i
\end{equation}
where $U_i$ is the membrane potential of the $i$th neuron. $C_{\text{m},i}$ is the membrane capacitance, $g_{\text{m},i}$ is the membrane conductance, and $E_{\text{r},i}$ is the neuron's resting potential. $I_i$ models an applied external bias current. For the $j$th synapse, $E_{ij}$ is the reversal potential, and $g_{ij}$ is the maximal synaptic conductance. The product of $g_{ij}$ and the activity of the $j$th presynaptic neuron $y_j$ represents the synaptic channel conductance. A monotonically increasing activation function $\phi$ can be used to define the relationship between the output $y_i$ and the state $U_i$. We set $\phi $ as a piecewise linear function $\mathbb{R} \to [0, 1]$:
\begin{equation}\label{eq:2}
  y_i = \phi (U_i) = \frac{\min \left(\max(U_i, E_\text{lo}), E_\text{hi}\right) - E_\text{lo}}{E_\text{hi} - E_\text{lo}}
\end{equation}
The parameters $E_\text{hi}$ and $E_\text{lo}$ are the upper and lower threshold of the activation function, respectively\footnote{Typical threshold values in an SNS network are  $E_\text{lo}=0$ and $E_\text{hi} = 20$}.

Although SNS networks have previously been simulated as robotic controllers based on the differential governing equation \cite{szczecinski2017functional,hunt_development_2017}, the continuous time representation makes comparisons with existing classical artificial neural network (ANN) approaches and implementation in machine learning frameworks challenging. To address this challenge, we can obtain a discretized version of the SNS model by applying the semi-implicit Euler method \cite{webster2020control,hasani2021liquid} to (\ref{eq:1}):
\begin{IEEEeqnarray}{rCl}\label{eq:3}
  C_{\text{m},i}\frac{U_i(t)-U_i(t-1)}{\Delta} &=& g_{\text{m},i}(E_{\text{r},i} - U_i(t)) + I_i\\
  && + \sum\limits_{j} g_{ij}\phi\left(U_j(t-1)\right) \left(E_{ij} - U_i(t)\right) \nonumber
\end{IEEEeqnarray}
where $t$ is the current time and $\Delta$ is the time step. Letting $\tau_i = C_{\text{m},i}/g_{\text{m},i}$, $w_{ij} = g_{ij}E_{ij}/g_{\text{m},i}$, $v_{ij} = g_{ij}/g_{\text{m},i}$, $b_i = I_i/g_{\text{m},i}$ and moving all terms containing $t$ to the left hand side, we can further simplify (\ref{eq:3}) to
\begin{IEEEeqnarray}{rCl}
  \hat{\boldsymbol\tau}_{t} &=& \frac{\boldsymbol\tau}{1 +\mathbf{V}\phi\left(\mathbf{h}_{t-1}\right)}\label{eq:4}\\
  \mathbf{z}_{t} &=& \frac{\Delta}{\hat{\boldsymbol\tau}_{t}+\Delta} \\\label{eq:5}
  \hat{\mathbf{h}}_{t} &=& \frac{\mathbf{b} + \mathbf{W}\phi\left(\mathbf{h}_{t-1}\right)}{1 +\mathbf{V}\phi\left(\mathbf{h}_{t-1}\right)} \label{eq:6}\\
  \mathbf{h}_{t} &=& (1 - \mathbf{z}_{t})\odot\mathbf{h}_{t-1}+ \mathbf{z}_{t}\odot\hat{\mathbf{h}}_{t}\label{eq:7}
\end{IEEEeqnarray}
where $\mathbf{h}_t$ denotes the state vector $[U_{1}(t), \cdots, U_n(t)]^T$, $\boldsymbol{\tau}$ and $\mathbf{b}$ denote the time constant vector $[\tau_{1}, \cdots, \tau_n]^T$ and bias vector $[b_{1}, \cdots, b_n]^T$, respectively. The weight $\mathbf{W}$ and $\mathbf{V}$ satisfy $\mathbf{W}_{ij} = w_{ij}$ and $\mathbf{V}_{ij} = v_{ij}$, respectively. The operator $\odot$ denotes the Hadamard product (element-wise product).

Equations (\ref{eq:4})-(\ref{eq:7}) imply that some classical ANNs are special cases of the SNS model. For example, when $\mathbf{V} = \boldsymbol{0}$, (\ref{eq:4})-(\ref{eq:7}) degenerate to the semi-implicit Euler discretization of the continuous recurrent neural network (CTRNN) model \cite{FUNAHASHI1993801}
\begin{IEEEeqnarray}{rCl}
  \hat{\mathbf{h}}_{t} &=& \mathbf{b} + \mathbf{W}\phi\left(\mathbf{h}_{t-1}\right) \label{eq:8}\\
  \mathbf{h}_{t} &=& \frac{\boldsymbol{\tau}}{\boldsymbol\tau +\Delta}\odot\mathbf{h}_{t-1}+ \frac{\Delta}{\boldsymbol\tau +\Delta}\odot\hat{\mathbf{h}}_{t}\label{eq:9}
\end{IEEEeqnarray}
If we further set $\boldsymbol\tau = \boldsymbol{0}$, then (\ref{eq:8})-(\ref{eq:9}) becomes the Vanilla recurrent neural network (Vanilla RNN), or a discretized version of the neural ordinary differential equations \cite{NEURIPS2018_69386f6b}:
\begin{equation}\label{eq:10}
  \mathbf{h}_{t} = \mathbf{b} + \mathbf{W}\phi\left(\mathbf{h}_{t-1}\right)
\end{equation}
When we study constant time series $\mathbf{h}_t = \mathbf{h}_{t-1}$, the time $t$ is irrelevant and (\ref{eq:10}) is equivalent to a multilayer perceptron (MLP). Note that (\ref{eq:4})-(\ref{eq:7}) have a similar form to the gated recurrent unit (GRU) \cite{DBLP:journals/corr/ChoMBB14, NEURIPS2021_05546b0e}.

The network parameters to learn in SNSs are $\mathbf{W}$, $\mathbf{V}$, $\boldsymbol\tau$ ,and $\mathbf{b}$ in (\ref{eq:4})-(\ref{eq:7}). The discretized version of SNSs can be trained by classical methods in RNN, such as backpropagation through time (BPTT) \cite{BPTT} for time series prediction problems. 

\subsection{Learning Nonlinear Operations with Discretized SNSs}\label{OperationLearning}

Each conductance-based synapse in the SNS is parameterized by two parameters ($g_{ij}$, $E_{ij}$ in (\ref{eq:1}) or $\mathbf{W}$, $\mathbf{V}$ in (\ref{eq:4})-(\ref{eq:7}) that can be related to biological neural network measurements. In contrast, a synapse in ANN models is parameterized by a single parameter (synaptic weight $\mathbf{W}$ in (\ref{eq:8}) and (\ref{eq:10})). The additional weight matrix $\mathbf{V}$ in the SNS allows the inputs to modify the apparent neuron time constants ($\hat{\boldsymbol\tau}_t$ in (\ref{eq:4})) \cite{lechner2019designing,lechner2020neural,hasani2021liquid}. This mechanism is similar to update gates in GRU \cite{DBLP:journals/corr/ChoMBB14}, which indicates the SNS has potential to learn time series dependencies. The incorporation of $\mathbf{V}$ turns the weighted sum term in (\ref{eq:8}) into a fractional function in (\ref{eq:6}), enabling the SNS to capture learning phenomena inspired by more complex dendritic integration \cite{10.1093/oso/9780195104912.001.0001, dendritic_computation_1,koch2000role,groschner2022biophysical}. Such nonlinear interactions have potential for integrating multimodal information or performing conditional computation in robotic control \cite{DBLP:conf/iclr/JayakumarCMSROT20}.
\begin{figure*}[tb]
 \centering
 \includegraphics[width=0.8\textwidth]{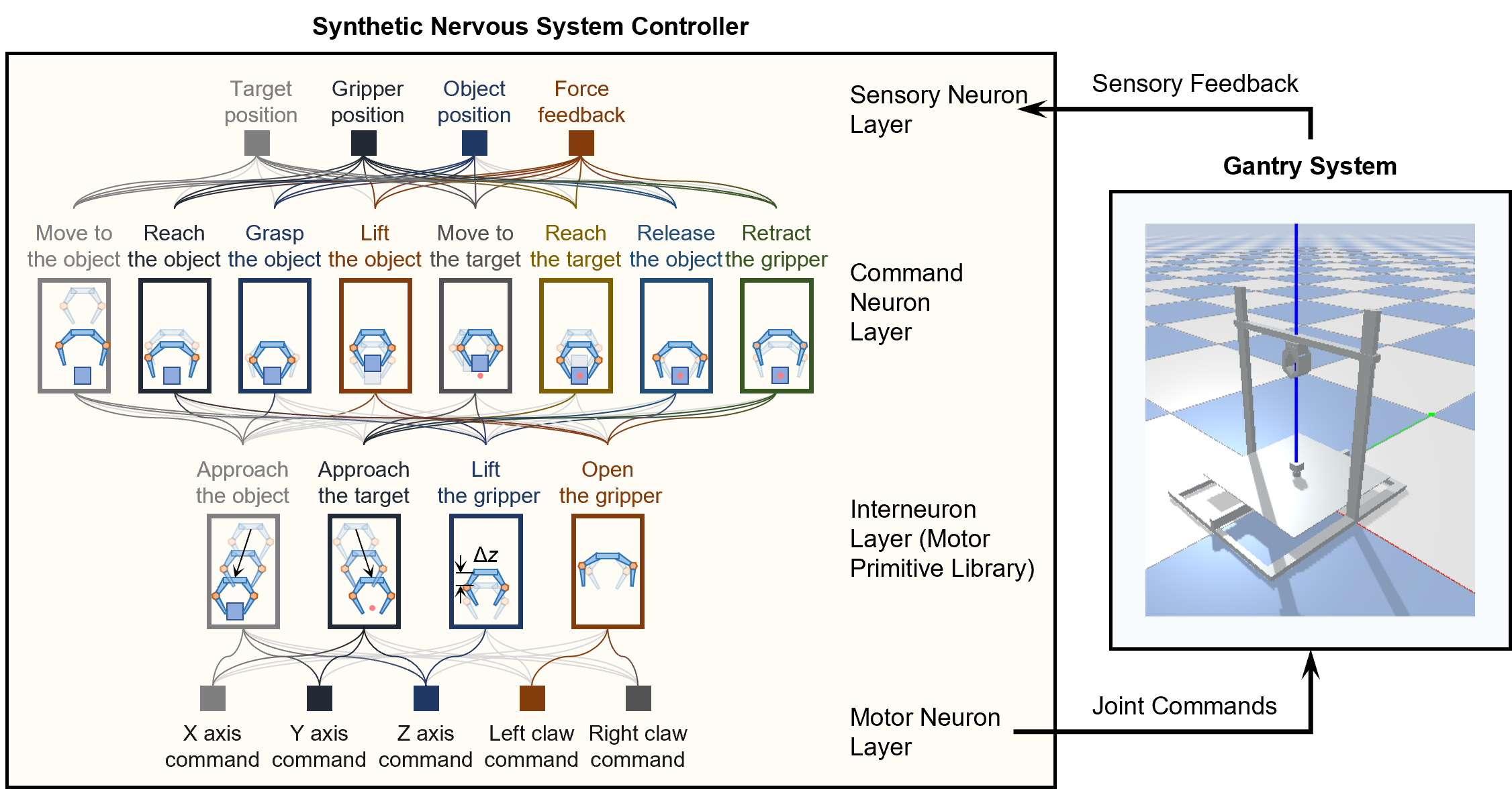}
 \vspace{-10pt}
 \caption{Architecture for the SNS-based gantry control system. Each subnetwork in the command neuron layer and interneuron layer represents a neuron in the SNS controller. The command neuron layer contains 8 subnetworks whose activations represent eight different high-level commands for pick-and-place control. Each high-level command can be decomposed into subtasks existing in the motor primitive library. For example, the lifting the object command can be decomposed into approaching the object and lifting up the closed gripper. Similarly, each subnetwork in the interneuron layer represents a specific motor primitive which can be further decomposed into corresponding patterns of motor neuron activities. Therefore, subnetworks in the command neuron layer and the interneuron layer only need sparse connections (represented by colored lines) to selectively activate or inhibit related subnetworks in the lower layer. We grayed out the connections existing in the fully connected neural network but not existing in our architecture to emphasize the sparsity. The sensory neuron layer is responsible for collecting the sensory feedback and sequentially activating the subnetworks in the command neuron layer.}
 \label{fig:sysArch}
 \vspace{-15pt}
\end{figure*}

To see how the rational dendritic integration term in the SNS increases the ability to perform such foundational calculations, we can consider a simple SNS with three neurons $\mathbf{h}_t = [U_{\text{pre},1},\, U_{\text{pre},2},\, U_{\text{post}}]$ \cite{szczecinski2017functional}. If we set $E_\text{lo}=0$, $E_\text{hi}=20$ and assume $ U_{\text{pre},1}$ and $U_{\text{pre},2}$ are not saturated, then $\phi (U_{\text{pre},1}) = U_{\text{pre},1}/20$, $\phi (U_{\text{pre},2}) = U_{\text{pre},2}/20$. If $\mathbf{b} = \boldsymbol{0}$, $\boldsymbol\tau = \boldsymbol{0}$, $\mathbf{W}_{3,\ast} = [20,\, 0,\, 0]$, and $\mathbf{V}_{3,\ast} = [0,\, 20,\, 0]$, we have\footnote{The Asterisks notation represents a specific row in the matrix}
\begin{equation}
  U_{\text{post}}(t) = \frac{U_{\text{pre},1}(t-1)}{1 +U_{\text{pre},2}(t-1)}
\end{equation}
When $U_{\text{pre},2} \gg 1$, this network approximates division. The effect of $U_{\text{pre},2}$ is similar to shunting inhibition in computational neuroscience. Shunting inhibition is key to neuromodulation \cite{10.1093/oso/9780195104912.001.0001}, but it can not be easily achieved in classical ANNs.




We hypothesized that compact and sparse SNS networks can learn parameters for arithmetic. To test this hypothesis, we compared the ability of our SNSs and compact multilayer perceptrons (MLPs), a classical ANN model, to learn four operations: addition, subtraction, division, and multiplication. The SNS for each operation is inspired by previous work on functional subnetworks \cite{szczecinski2017functional}. We adopted similar architecture for MLPs (single layer, layer size 1 for addition, subtraction, and division; for multiplication, two layers, layer size (2,1)). We formulated the arithmetic operation learning as a time sequence regression task and adopted the mean square error as the training loss function. The training set includes 1000 training examples with two constant feature series and the corresponding label series. We set $\Delta = 0.1 \text{sec}$ and the sequence length as 50. The values of the constant feature series are randomly selected from $[E_\text{lo},\, E_\text{hi}]$. We used Adam \cite{Adam} as the optimizer. To compare learning performance between SNS and MLP networks, we compared training loss.

\subsection{Designing and Learning Controllers with Discretized SNSs: Pick-and-Place Case Study}
\begin{figure*}[ht]
 \centering
 \includegraphics[width=0.94\textwidth]{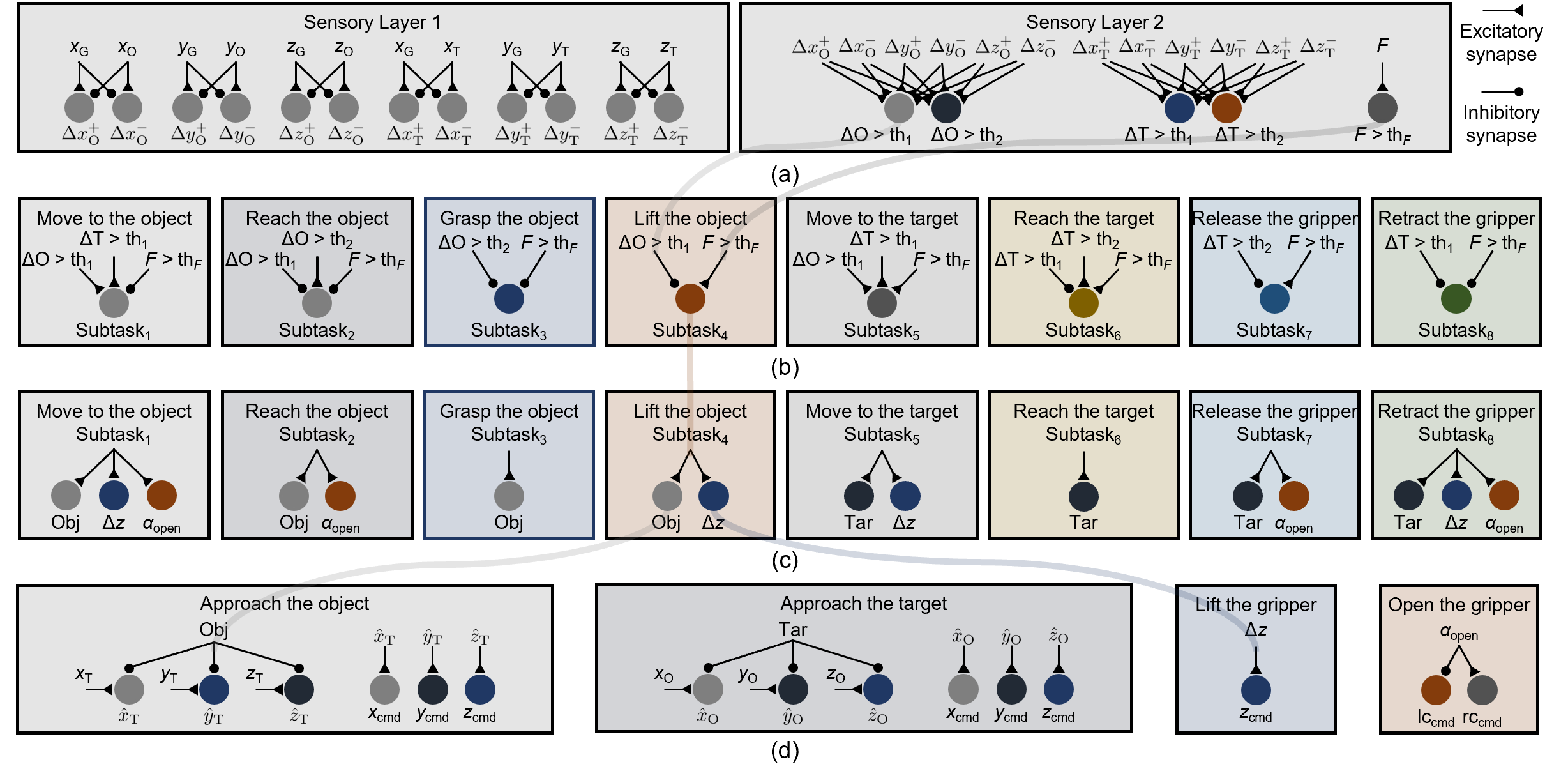}
 \vspace{-12pt}
 \caption{The hierarchical SNS controller for pick-and-place control. (a) The first sensory layer calculates the difference in position between the object and the gripper ($\Delta x_\text{O},\Delta y_\text{O},\Delta z_\text{O}$), and between the target and the gripper ($\Delta x_\text{T},\Delta y_\text{T},\Delta z_\text{T}$) in the $x,y,z$ axes. The second sensory layer receives inputs from the first layer and from force feedback to determine whether the distance is within a user-specified range bounded by ($\text{th}_\text{1}$) and ($\text{th}_{\text{2}}$), and whether the force is greater than a threshold ($\text{th}_F$). (b) The command layer contains 8 neurons ($\text{Subtask}_{1 - 8}$) corresponding to 8 subtasks to accomplish. Neurons in this layer are activated if all presynaptic neurons with excitatory synapses are firing while those with inhibitory synapses are silent. (c) The interneuron layer contains four neurons representing four motor primitives: Obj (moving to the object), Tar (moving to the target), $\Delta z$ (lifting up the gripper), and $\alpha_\text{open}$ (opening the gripper). Neurons in this layer are activated if any of the presynaptic neurons are firing. (d) The motor neurons layer contains motor neurons for $xyz$ movement ($xyz_\text{cmd}$) and the angles of the claws of the gripper, ($\text{lc}_\text{cmd}$ and $\text{rc}_\text{cmd}$). These neurons receive transmission pathways \cite{szczecinski2017functional} from their presynaptic neurons. We also include neurons receiving the object position ($\hat{x}_\text{O}$, $\hat{y}_\text{O}$, $\hat{z}_\text{O}$) and the target position ($\hat{x}_\text{T}$, $\hat{y}_\text{T}$, $\hat{z}_\text{T}$) in this layer. Neuron Obj has inhibitory modulation connections to $\hat{x}_\text{T}$, $\hat{y}_\text{T}$ and $\hat{z}_\text{T}$, while Tar has inhibitory modulation connections to $\hat{x}_\text{O}$, $\hat{y}_\text{O}$ and $\hat{z}_\text{O}$. The existence of these modulation pathways allows the SNS controller to select from moving to the object or moving to the target. The subnetworks highlighted by the blue borders help us walk through an example showing how a high-level command is triggered and implemented. If the gripper is close to the original position of the object ($\Delta\text{O}>\text{th}_1$ not activated) and the gripper has grasped the object ($F > \text{th}_F$ activated), the command neuron corresponding to lifting up the object ($\text{Subtask}_4$) starts firing. It in turn activates two interneurons Obj and $\Delta z$. Neuron Obj inhibits the sensing of the target position so that the motor neurons receive the original position of the object, while $\Delta z$ increases the activity of $z_\text{cmd}$ by a small amount. Therefore, the gripper can lift the grasped object.}
 \vspace{-10pt}
 \label{fig:SNScontroller}
\end{figure*}

To assess the ability of our discretized SNS approach to learn robotic control parameters, we developed a grasper pick-and-place controller. The network architecture is abstractly inspired by that of feeding in the sea slug \textit{Aplysia californica} \cite{webster2020control}. Generally, the neurons responsible for \textit{Aplysia} feeding can be organized into sensory neurons, motor neurons, interneurons, and command-like neurons. Command-like neurons receive feedback signals from sensory neurons. They control transitions between feeding behaviors by activating interneurons based on sensory feedback \cite{CBIs1,CBIS2}. The interneurons can be organized into subnetworks that realize essential control functions \cite{Interneurons}. In our architecture (Figure \ref{fig:sysArch}), sensory signals for the target position, gripper position, object position, and force feedback are received by a sensory layer. These neurons are connected to a command layer to control behavioral switching, which projects onto an interneuron layer that coordinates motor primitives and sends signals to the motor neuron layer for actuator control. For each function in the network architecture (Fig. \ref{fig:sysArch}), the SNS controller (Fig. \ref{fig:SNScontroller}) possesses a corresponding neuron expressed as the discretized SNS described in section \ref{DSNS}. Network parameters are initialized and trained by the supervised learning paradigm in Section \ref{OperationLearning} so that subnetworks can generate the specified input-output relationships without using analytical methods like FSNs. Fig. \ref{fig:SNScontroller} also demonstrates the synaptic polarity of connections found by the learning process.

\subsection{Simulation Environment for Gantry SNS Parameter Tuning}

\begin{figure}[htb!]
  \includegraphics[width=0.85\columnwidth]{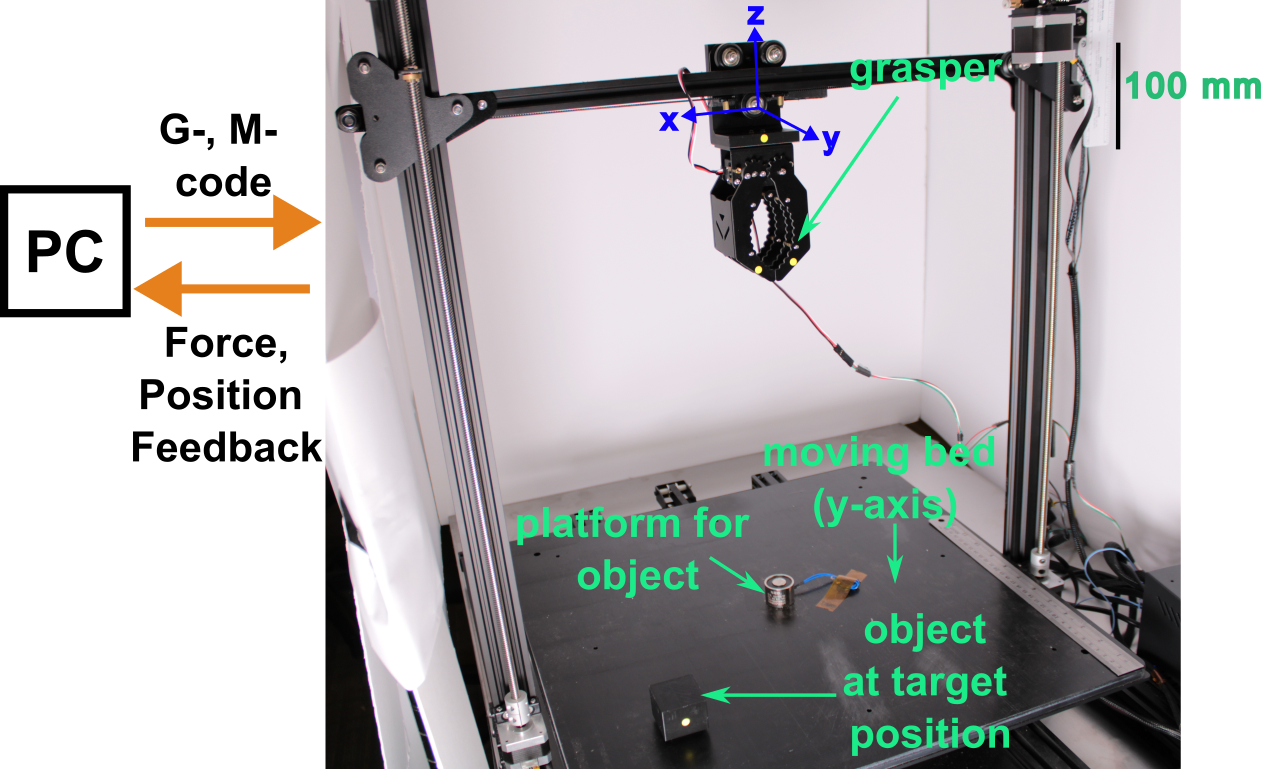}
  \vspace{-8pt}
  \caption{Grasping robot used to validate SNS control. System is in the origin position of $(0,0,0)$. Object is at the target position. The grasper is mounted such that it can move in the $x-z$ plane. The platform moves in the $y$ axis.}
  \vspace{-15pt}
  \label{fig:ExperimentalSetup}
  \centering
\end{figure}

To enable fine-tuning of the SNS before use in controlling a physical system (Fig.~\ref{fig:ExperimentalSetup}), a simulation environment was created in PyBullet (Fig.~\ref{fig:sysArch}) \cite{coumans2016pybullet}. This simulation environment captures the major features of the robot: ($1$) the dimensions of the gantry, grasper, and object are true-to-size, and $(2)$ position control, where at each timestep a position command of where the system should be at the next timestep was sent to the simulation. The velocity limits of the axes were tuned to correspond with the peak velocities of the physical system.

\vspace{-5pt}
\subsection{Experimental Validation}
A Creality CR-10 S5 3D printer \cite{CrealityCr10S5} was modified to accommodate a servomotor actuated grasper to validate SNS control in a physical system (Fig.~\ref{fig:ExperimentalSetup}). This system could translate in three orthogonal axes $(x,y,z)$, as well as control the angle between the claws of the symmetric grasper. The 3D printer's firmware was modified to control grasper state via M-code commands and query the 3D printer's stepper motor positions. Assuming no missed steps during actuation, this enabled pseudo-closed loop control of the gantry in the $x$, $y$, and $z$ axes, as well as the grasper angle, $\theta_{grasper}$. Contact force information was provided by manually pressing a tactile pushbutton switch when contact was made between the grasper and the object and was registered by the gantry firmware. 

The gantry position and force were inputs for the SNS (Fig.~\ref{fig:sysArch}). At each timestep, the SNS controller uses these inputs to produce the next set of $(x,y,z,\theta_{grasper})$ target coordinates. The object position input was set as $(0,0,-0.305)$~\si{m}, and the target position as $(0.15,0.15,-0.310)$~\si{m}. The object was a 19.7~\si{g} cube of size $(39, 39, 34.5)$~\si{mm}. 

\begin{figure*}[t]
 \centering
 \includegraphics[width=0.99\textwidth]{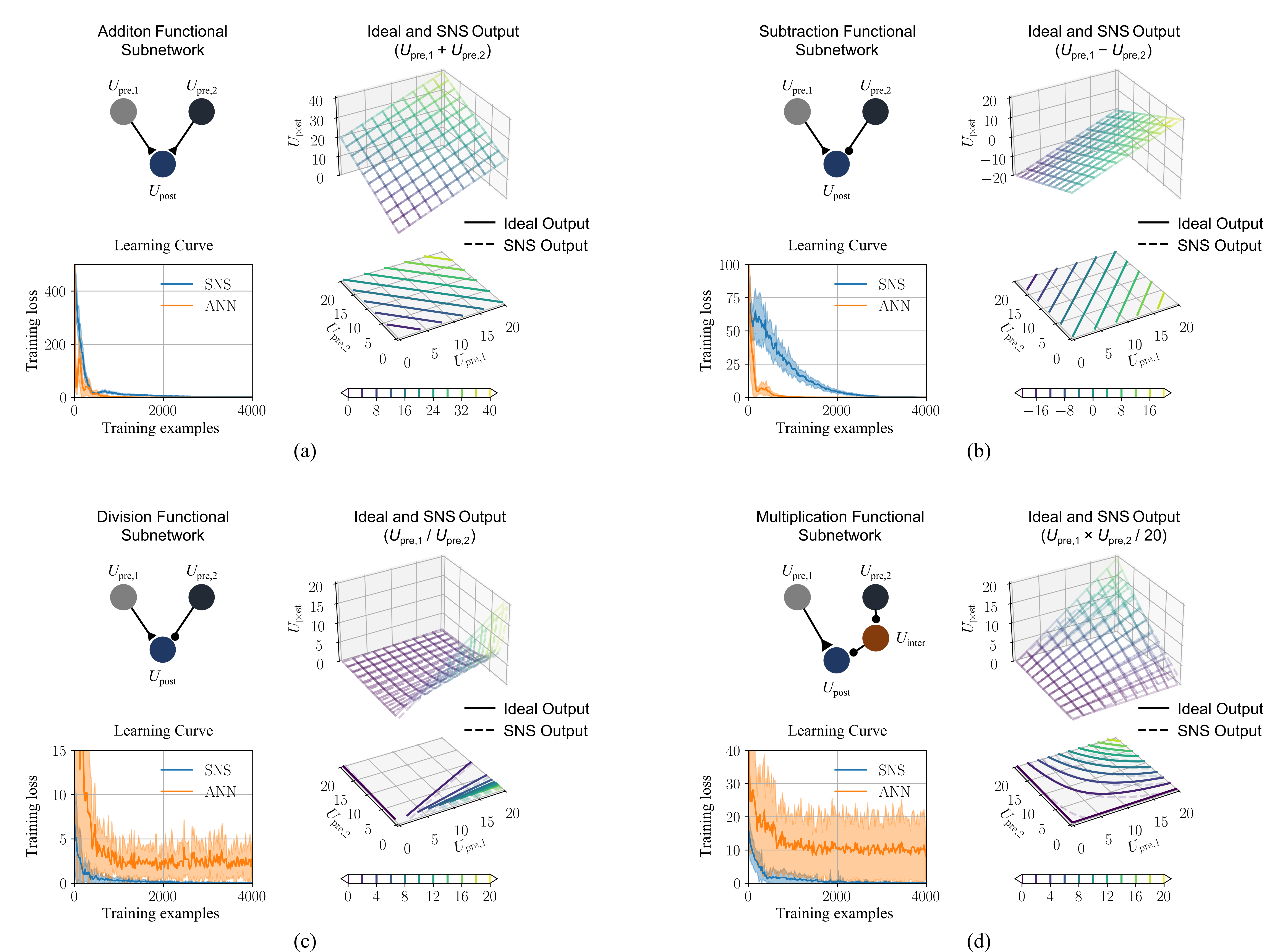}
 \vspace{-10pt}
 \caption{Network diagrams, learning curves, functions, and contours of the ideal and SNS output for addition (a), subtraction (b), division (c), and multiplication (d). Triangular synaptic terminations represent synapses becoming excitatory after learning, and filled round terminations represent synapses becoming inhibitory after learning. In the multiplication learning task, we initialized the connection between the presynaptic neuron to the interneurons as an inhibitory synapse to avoid potential local minima. All parameters are initialized randomly for other tasks. The comparison between the ideal and SNS output and their corresponding contours shows that SNSs can perform these operations. The learning curves also suggest SNSs can learn  multiplication and division quickly, while the compact ANNs cannot generate close approximations to these nonlinear operations.}
 \vspace{-10pt}
 \label{fig:arithmatic_learning}
\end{figure*}

\section{Results and Discussion}

\subsection{SNS Networks can Learn Parameters for Key Mathematical Calculations better than Compact MLPs}

Both the SNSs and compact MLPs specified in  section \ref{OperationLearning} could learn the arithmetic operations for addition and subtraction (Fig.\ref{fig:arithmatic_learning}(a-b)). The learning process of the MLPs is faster than the SNSs in these two cases because the operations to learn are linear functions, and the additional parameter $\mathbf{V}$ in SNSs is redundant. However, only the SNSs could learn close approximations to the division and multiplication operations (Fig.\ref{fig:arithmatic_learning}(c-d)). Compact MLPs are insufficient to learn them because they lack inductive bias for nonlinear computation.

\begin{figure*}[hbt!]
 \centering
 \includegraphics[width=0.98\textwidth]{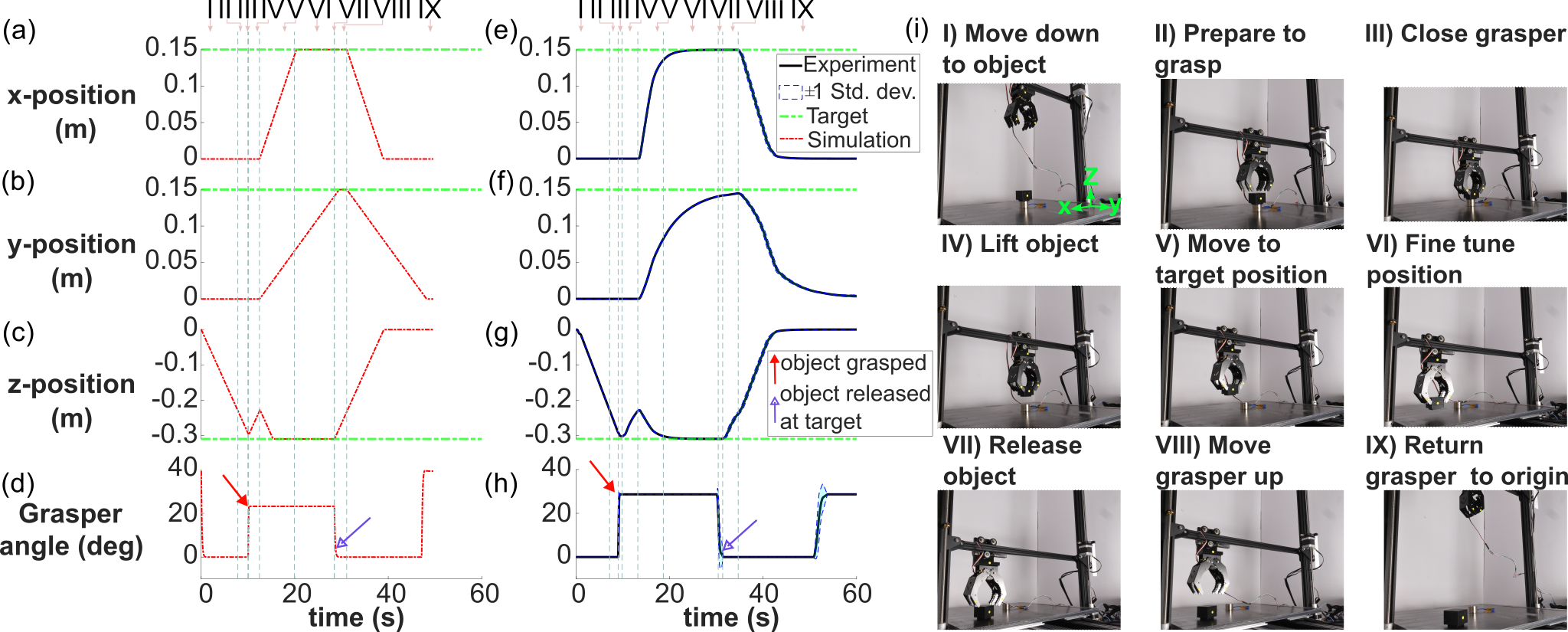}
 \caption{(a), (b), (c) Kinematics of the $x$, $y$, and $z$ axes, respectively, of the simulated gantry under SNS control. (d) Grasper angle kinematics. (e), (f), (g), (h) Kinematics of the $x$, $y$, $z$, and the grasper, respectively, of the real gantry under real-time SNS control.  (i) State of the gantry, grasper, and object under real-time control. States \textbf{I}) to \textbf{VIII}) correspond to the dominant Command Neuron active at that point in time (Fig.~\ref{fig:sysArch}). State \textbf{IX}) represents the return of the grasper to its original position. This behavior naturally comes about from the dynamics of states \textbf{I} to \textbf{XIII} by setting the object position to the origin.}
 \vspace{-10pt}
 \label{ExpVsSimResults}
\end{figure*}

\subsection{The SNS is Capable of Real-time Grasping Control}
The SNS-controlled real-time, real-world grasper system (Fig.~\ref{fig:ExperimentalSetup}) was able to successfully recreate the behavior of the SNS-controlled simulation, namely: ($1$) grasp the object, ($2$) move the object to the target position, and ($3$) return to the starting position (Fig.~\ref{ExpVsSimResults}). Importantly, the SNS controller whose parameters were tuned in the simulation environment was able to be used as-is in the control of the real grasper system without the need for any additional parameter tuning.   


In addition to the SNS computation of the next commanded state $(x,y,z,\theta_{grasper})$, the controller also had to handle the communication with the gantry system. These I/O operations, where commands and state information were exchanged between the PC and gantry system, resulted in a longer average controller timestep for the real gantry system ($85.5\pm 14.9$~\si{ms}) when compared to the simulation ($16.5 \pm 4.2$~\si{ms}). For both the real and simulated grasper systems, the average time taken to execute the SNS computations was comparable ($5.9 \pm 1.4$~\si{ms} and $4.8 \pm 2.2$~\si{ms}, respectively). Future work will explore avenues for further improvement of the controller bandwidth by optimizing the computations performed in the SNS and in the I/O operations between the controller and grasper system.

It should be noted that the gantry essentially operates under position control as stepper motors are used to drive the $x$, $y$, and $z$ axes. Hence, the position of the gantry could be directly controlled without considering the forces needed to produce such a motion, as long as the axes' peak acceleration and velocity limits were respected. Future work will explore the ability of the SNS to control robotic systems with more complicated dynamics, such as a 6-DOF robotic arm. 
\vspace{-3pt}
\subsection{Kinematics of the Real-World Grasper System are Similar to those of the Simulation under SNS Control}
\vspace{-3pt}
Despite the increased controller timestep in the physical grasper system, the kinematics of the physical SNS-controlled grasper (Fig.~\ref{ExpVsSimResults}e-h) are qualitatively similar to those of the simulated grasper system (Fig.~\ref{ExpVsSimResults}a-d). Moreover, the physical system was capable of executing the behavioral patterns prescribed by the SNS's Command Neuron Layer (Fig.~\ref{fig:sysArch}), which allowed the grasper to replicate the simulation's success in grasping and moving the object to the target location (Fig.~\ref{ExpVsSimResults}i). 

However, there were notable differences in the kinematics of the real-world physical system when compared to the simulation. In phases \textbf{III} to \textbf{IV} (Fig.~\ref{ExpVsSimResults}i), the gantry grasps the object and begins to ascend vertically, corresponding to the ``Grasp Object" and ``Lift Object" Command Neurons, respectively (Fig.~\ref{fig:sysArch}). In the simulation, the transition from the descent of the grasper in the $z$ axis to the ascent of the grasper occurs instantaneously as contact information can be determined immediately (Fig.~\ref{ExpVsSimResults}c). In the physical grasper system, however, there is a small delay between these actions (Fig.~\ref{ExpVsSimResults}g), as contact is determined by a manual press of a push-button. This is also the reason for the small delay between the release of the object at the end of phase \textbf{VII} (Fig.~\ref{ExpVsSimResults}h) and the beginning of the grasper retraction in phase \textbf{VIII} (Fig.~\ref{ExpVsSimResults}g). The simulation also maintained higher velocities in the $y$ and $+z$ axes of the physical system. Possible reasons for the slower robot motions in those axes are that the $y$ axis has the largest mass of all three axes, and the $+z$ axis must work against gravity. Further tuning of the simulation's dynamics may allow for a better match to the physical system. 

Though not evident from Fig.~\ref{ExpVsSimResults}e-h, the physical grasper system would intermittently exhibit non-smooth movement. This was most acute in the $x$-$y$ plane during phases \textbf{V} to \textbf{VIII} when the difference between the commanded position and the current position was small because the gantry slowly approached the target position to release the object. The choppy motion observed during this behavior can be attributed to the longer than expected controller timestep and the trapezoidal velocity motion profile of the axes as they moved towards the commanded position. If the time taken to move from the current position to the commanded position was smaller than the time to execute the control loop, then such choppy motion could occur. To compensate for the increased overhead introduced by the I/O operations, moves were buffered on the controller side (maximum of 10 move commands) to improve motion smoothness. Future experiments will benefit from improving the controller timestep and optimizing the motion profile to prevent short-burst decelerations and accelerations.

\section{Conclusions and Future Work}
\vspace{-5pt}

Bioinspired approaches to the control of robots have the potential to enable the type of behavioral flexibility found in animals.  In this paper, we have shown that bioinspired Synthetic Nervous Systems (SNSs) are a promising approach to robotic control as they have reduced training time for learning foundational mathematical operations in robotic control, such as division and multiplication, when compared to Vanilla ANNs (Fig.~\ref{fig:arithmatic_learning}). We have also shown that SNSs can successfully control a robotic grasper to perform a pick-and-place manipulation task.  We demonstrated that an SNS controller can be tuned in a simulation environment and subsequently used in a real physical system for a pick-and-place task without additional tuning (Fig.~\ref{ExpVsSimResults}). The real grasper system's kinematics closely followed the simulation's behavior. 

Ultimately, the use of supervised learning to determine SNS parameters (Fig.~\ref{fig:arithmatic_learning}), and the hierarchical technique to create a pick-and-place controller (Fig.~\ref{fig:SNScontroller}) described in this paper can be extended to other robotic systems to leverage the SNS's capability of performing foundational operations efficiently.  Future work will validate the SNS real-time controller for use in controlling robotic systems with more complex dynamics, such as a 6-DOF robotic arm, or more complex perception, such as proprioceptive and haptic feedback to manipulate soft, fragile, and irregularly shaped objects.  As the SNS includes elements of real neural dynamics, we will extend this bioinspired controller to include plasticity to realize real-time learning and adaptation \cite{stepanyants_geometry_2002,morton_neural_1994} which can automatically find a hierarchical and sparse controller for manipulation tasks.

\vspace{-5pt}
\section*{Acknowledgment}
\vspace{-5pt}
We thank Kevin Dai for help in proofreading the manuscript. Funding to attend this conference was provided by the CMU GSA/Provost Conference Funding.


\clearpage

\bibliography{bibtex/bib/IEEEabrv.bib,bibtex/bib/IEEEexample.bib}{}
\bibliographystyle{IEEEtran}

\end{document}